\documentclass[nohyperref]{article}

\usepackage{microtype}
\usepackage{booktabs} 

\usepackage{hyperref}

\usepackage[accepted]{icml2022}

\usepackage{amsmath}
\usepackage{amssymb}
\usepackage{mathtools}
\usepackage{amsthm}

\usepackage{amsmath,amsfonts,bm}

\def\eqref#1{equation~\ref{#1}}

\def\1{\bm{1}}

\DeclareMathAlphabet{\mathsfit}{\encodingdefault}{\sfdefault}{m}{sl}
\SetMathAlphabet{\mathsfit}{bold}{\encodingdefault}{\sfdefault}{bx}{n}

\newcommand{\R}{\mathbb{R}}

\usepackage{graphicx}

\usepackage[utf8]{inputenc} 
\usepackage[T1]{fontenc}    
\usepackage{url}            

\usepackage{amsfonts}       
\usepackage{nicefrac}       
\usepackage{microtype}      
\usepackage{subcaption}
\usepackage{comment}
\usepackage{graphicx}
\usepackage{multirow}
\usepackage{bm}
\usepackage{pifont}
\usepackage{mathtools}

\usepackage[capitalize,noabbrev]{cleveref}

\theoremstyle{plain}

\theoremstyle{definition}

\theoremstyle{remark}

\usepackage[textsize=tiny]{todonotes}

\icmltitlerunning{Distribution-aware Goal Prediction and Conformant Model-based Planning for Safe Autonomous Driving}

\newcommand{\cut}[1]{}
\newcommand{\para}[1]{{\noindent\textbf{#1}}}
\newcommand{\parai}[1]{{\noindent\textit{#1}}}
\newcommand{\ourmodel}{\texttt{DGP}}
\newcommand{\frenet}{Fren\'et}

\newcommand{\jf}[1]{\textcolor{red}{[\textbf{JF:} #1]}}

\newcommand{\grey}[1]{\textcolor{gray}{#1}}

\newcommand{\parencite}[1]{\citep{#1}}%
\newcommand{\textcite}[1]{\citet{#1}}%

\begin{document}

\twocolumn[
\icmltitle{Distribution-aware Goal Prediction and Conformant Model-based Planning \\for Safe Autonomous Driving}



\icmlsetsymbol{equal}{*}

\begin{icmlauthorlist}
\icmlauthor{Jonathan Francis}{equal,scs,bosch}
\icmlauthor{Bingqing Chen}{equal,scs}
\icmlauthor{Weiran Yao}{equal,scs}
\icmlauthor{Eric Nyberg}{scs}
\icmlauthor{Jean Oh}{scs}
\end{icmlauthorlist}

\icmlaffiliation{scs}{School of Computer Science, Carnegie Mellon University; Pittsburgh, Pennsylvania, United States}
\icmlaffiliation{bosch}{Bosch Center for Artificial Intelligence; Pittsburgh, Pennsylvania, United States}

\icmlcorrespondingauthor{Jonathan Francis}{jon.francis@us.bosch.com}

\icmlkeywords{Machine Learning, ICML}

\vskip 0.3in
]



\printAffiliationsAndNotice{\icmlEqualContribution} 

\begin{abstract}
The feasibility of collecting a large amount of expert demonstrations has inspired growing research interests in learning-to-drive settings, where models learn by imitating the driving behaviour from experts. However, exclusively relying on imitation can limit agents' generalisability to novel scenarios that are outside the support of the training data. In this paper, we address this challenge by factorising the driving task, based on the intuition that modular architectures are more generalisable and more robust to changes in the environment compared to monolithic, end-to-end frameworks. Specifically, we draw inspiration from the trajectory forecasting community and reformulate the learning-to-drive task as obstacle-aware perception and grounding, distribution-aware goal prediction, and model-based planning. Firstly, we train the obstacle-aware perception module to extract salient representation of the visual context. Then, we learn a multi-modal goal distribution by performing conditional density-estimation using normalising flow. Finally, we ground candidate trajectory predictions road geometry, and plan the actions based on  on vehicle dynamics. Under the CARLA simulator, we report state-of-the-art results on the CARNOVEL benchmark.
\end{abstract}

\section{Introduction}
\label{section:intro}

\begin{figure*}[ht]
\footnotesize
\centering
\includegraphics[width=\linewidth]{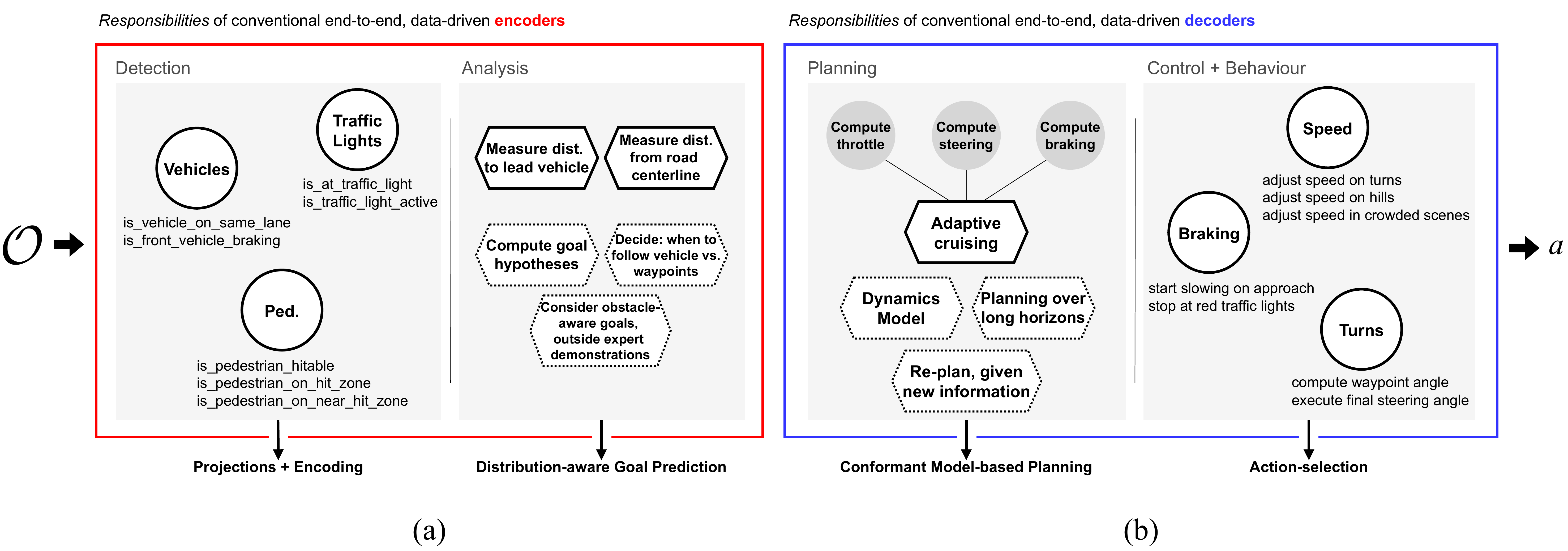}
\caption{Issues with end-to-end imitative pipelines. The (a) {\color{red} \textbf{red}} and (b) {\color{blue} \textbf{blue}} boxes illustrate the scope of responsibilities of conventional data-driven encoders and decoders, respectively, in the overall pursuit of replicating human driving behaviour. These include obstacle detection and scene analysis, and planning and control. Entities with dotted lines indicate behavioural components that lie outside the support of the expert demonstrations in typical learning-to-drive AI tasks, as in CARLA simulation, such as: computing goal alternatives, in response to dynamic obstacle behaviour or re-planning over long horizons when presented with new route information. As a result, it is neither possible for end-to-end imitative models to recover these skills from data, nor for them to exhibit the necessary degree of internal specialisation, without role assignment through the adoption of modular training. Our modular decomposition scheme (bottom arrows) is motivated by this taxonomy, as well as by the shortcomings of alternative decomposition schemes.}
\label{ad:l2d:fig:issue_with_n2n_imitation}
\end{figure*} 


Achieving generalisability to novel scenarios in urban autonomous driving remains a challenging task for artificial intelligence (AI). Recent approaches have shown promising results in end-to-end imitation learning from expert demonstrations, wherein agents learn policies that replicate the experts' actions, at each time-step, given the corresponding observations \parencite{bojarski2016end, codevilla2018end, codevilla2019exploring, muller2006off, pan2017agile, pomerleau1989alvinn, ohn2020learning, chen2020learning}. Despite this progress, end-to-end imitative models often cannot capture the causal structures that underlie expert-environment interactions, leading models to misidentify the correct mappings from the observations \parencite{dehaan2019causal}. Furthermore, if the coverage of expert demonstrations does not extend to \textit{all} scenarios that the agent will encounter during test time, the agent will generate spurious actions in response to these out-of-distribution (OOD) observations \parencite{filos2020can}.

In an effort to tackle \textit{part} of this issue, recent works deviate from the end-to-end learning paradigm in their respective problem domains, opting instead to decompose learning into sub-modules, for trajectory forecasting \parencite{rhinehart2018deep, filos2020can}, indoor robot navigation \parencite{chen2020setwaypts}, learnable robot exploration \parencite{chaplot2020learning, chen2019exploration}, and autonomous energy systems \parencite{chen2020citylearn, chen2020cohort}. Here, the intuition is that, by breaking down the inference problem into smaller units, more control over the inference step is obtained and the causal misidentification issue is somewhat avoided, by factorising the problems via engineering judgement. 
The modularity of those approaches resonates with the proposed method, however those approaches use only the classical global-local hierarchical planning paradigm, where the responsibility of performing feature-extraction while also attempting to (implicitly) model the environment dynamics is still contained within a single unit, leading to spurious predictions in unseen environments. We proceed a step further, by defining modules in the learning-to-drive setting, such that each module's task is directly attributable to the components of behaviour expected of an expert agent (see Figure \ref{ad:l2d:fig:issue_with_n2n_imitation}). In our decomposition, modules are given specialised roles (e.g., obstacle-awareness, goal/intention-prediction, and explicitly modelling environmental dynamics), improving the tractability of their respective sub-tasks and their complementarity towards the downstream objective.

However, the challenge of generalisability still remains. How can we effectively utilise the expert's prior experience (e.g., in the form of expert demonstrations), while also achieving generalisability to novel scenarios? Some recent works from the trajectory forecasting community formulate a dual-objective optimisation, coupling an imitation objective with a goal likelihood term \parencite{rhinehart2018r2p2, rhinehart2019precog, park2020diverse, bhat2020trajformer}, arguing that the two ideals of using prior expert experience and generalising can be unified. A common issue with this formulation is that models are incentivised to \textit{trade-off} the two objectives, rather than inherit their individual benefits. Samples from the goal distribution may not be sufficiently diverse, if the expert demonstrations did not provide sufficient coverage over the modes in the distribution over all possible predictions. Furthermore, predictions may not be admissible, discussed by \textcite{park2020diverse, francis2021core}, without some bias to adhere to, e.g., known physical constraints, as in \textit{Verlet integration} \parencite{verlet1967computer}. In this work, we utilise expert demonstrations for pre-training sub-modules and for density estimation, but we also ground predictions on a differentiable vehicle kinematics model and we constrain predictions to respect road admissibility through geometrical projection of goal prediction. 

As a summary of our contributions, we produce a framework for generating diverse multi-mode predictions, for the learning-to-drive setting, that achieves improved generalisability through modular task structures, more informed goal likelihood density-estimation, and explicit grounding on differentiable vehicle kinematics for trajectory generation\cut{, and learnable trajectory-pruning through adversarial filtering and policy refinement}. Our approach is summarised in Figure \ref{ad:l2d:fig:main_figure}. First, (i) we define modular primitives, based on insights about the decomposable nature of human driving behaviour. Next, (ii) we pursue model generalisability by coupling an imitation prior objective with a goal likelihood term, enabling the agent to leverage expert knowledge, while modelling more diverse modes in the underlying distribution over all goal futures. Next, (iii) we ground candidate trajectories on vehicle kinematics\cut{, while learning to prune the predictions that are spurious}. Finally, under CARLA simulation, (iv) we report new state-of-the-art results on the CARNOVEL benchmark\cut{, while matching perfect performance on NoCrash and the original CARLA benchmarks}.

\section{Related Work}

In this section, we describe prior art that is closely related to the core attributes of our approach.

\para{Learning to drive.} \textcite{pomerleau1989alvinn} pioneered investigation of end-to-end imitation learning, for sensorimotor navigation in autonomous driving. Following some extensions \parencite{muller2006off, silver2010learning, bojarski2016end} with applications in lane-following, highway driving, and obstacle avoidance, more recent works adapted the classic imitative modelling approach to urban driving scenarios \parencite{codevilla2018end, bansal2018chauffeurnet, codevilla2019exploring, pan2017agile, ohn2020learning, chen2020learning}, with more complex road layouts and challenging dynamic obstacle interactions. Whereas the increased sample-efficiency from imitation allays much serious consideration of alternative learning paradigms, e.g., reinforcement, a common issue with imitative modelling arises from having to learn a representation from high-dimensional visual inputs, in highly-varying environments: even with sufficient data, models struggle to extract meaningful features from the input that are not confounded by high-frequency, label-independent variation (e.g., varied vehicle shapes, sensor miscalibration, different weather conditions, shadows, poor expert behaviour) \parencite{bansal2018chauffeurnet}. In fact, access to more samples can actually yield worse performance, as low-quality data can lead the model to misidentify basic causal structures, underlying expert-environment interactions \parencite{dehaan2019causal}. Following \textcite{codevilla2018end, codevilla2019exploring},  \textcite{ohn2020learning, chen2020learning, sauer2018conditional} use conditioning strategies, such as command variables, teacher networks, and mixtures of expert policies, in attempts to learn better conditional representations and thus reduce the search space for generating actions. However, a limitation of these works is that the number of modes that can be represented by these methods is limited by the number of pre-specified commands or experts---thereby limiting the model's generalisability to novel driving scenes.

\para{Control strategies for autonomous vehicles.} Aside from learning (e.g., neural) mappings from observations to actions, various works advocate for the use of feedback or model-based control: to simplify the learning process for the data-driven components of the framework, to replace the data-driven components entirely, or to ground neural predictions with explicit physical constraints. \textcite{chen2020learning} utilise a proportional-integral-derivative (PID) controller to track the agent's target velocity, while \textcite{sauer2018conditional} use a PID controller for longitudinal tracking and a Stanley Controller (SC) \parencite{thrun2006stanley} for lateral tracking, with respect to the road centerline. A feedback controller \textit{myopically} and \textit{reactively} determines its control actions based on deviations from the setpoints, whereas model-based controllers, such a model-predictive controller (MPC), can plan trajectories over long planning horizons by unrolling its model of the system dynamics. \textcite{kabzan2019learning, herman2021learn, chen2021safe, francis2022l2rarcv} implement MPC controllers for their autonomous racing tasks, using ground-truth vehicle states. 
In this work, we integrate our perception and goal-prediction modules with an MPC, in the context of urban driving, enabling our system to generate trajectories that conform to vehicle kinematics.

\para{Trajectory forecasting for autonomous driving.} 
The notion of characterising distributions over all possible agent predictions has seen exciting growth in the domain of trajectory forecasting for autonomous driving \parencite{lee2017desire, rhinehart2018r2p2, rhinehart2018deep, rhinehart2019precog, park2020diverse, filos2020can}. Whereas \textcite{lee2017desire} use past trajectories and scene context as input for predicting future trajectories, and they score the `goodness' of a trajectory as a learnable module, their method does not attempt to model the agent's predictive intent, e.g., as modes in a likelihood density. \textcite{rhinehart2018deep} incorporated the concept of \textit{goal-likelihood} into their Deep Imitative Model (DIM), and characterised the agent's objective via pre-specified geometric primitives: points, piece-wise linear segments, and polygons. However, their goal-likelihood is defined as simple set membership (i.e., within the pre-specified geometry or not). Intuitively, set membership is neither a necessary or sufficient condition for good driving behaviour (e.g., banking vs following waypoints; avoiding obstacles vs staying on waypoints; staying within drivable area vs. driving safely). \textcite{filos2020can} aimed to improve on DIM by evaluating the trajectory on the basis of an ensemble of expert likelihoods; while that gives a more robust estimate of the `goodness' of a trajectory, it neither considers dynamic obstacles nor more informative goal priors. Whereas multi-agent trajectory forecasting has slightly different intentions and implications than action planning in learning-to-drive settings, we nonetheless draw inspiration from the trajectory forecasting literature for their distributional interpretation of the ego-agent's intent, which we combine with information about the scene context, for improved obstacle-awareness in those predicted trajectories.

\section{Problem Definition}

We define, here, the terminology that we will use to characterise our problem. The ego-agent is a dynamic, on-road entity whose state is characterised by a 4D pose: the spatial position (consisting of $x$, and $y$ in a Cartesian world coordinate frame), the speed $v$, and the yaw angle $\theta$, which evolves over time. For the position of the ego-agent at control time-step $k$, we use the notation $\mathbf{x}_k = [x_k, y_k, v_k, \theta_k] \in \R^4$; for the agent's sequence of positions, from time-step $k_1$ to $k_2$, we use $\mathbf{x}_{k_1 : k_2}$. For the full sequence of the ego-agent's positions, for a single episode in the training data, we use (bold) $\bm{X}$. Setting $k_0$ as the present state, we define the agent's historical trajectory $t\leq k_0$ to be $\bm{X}_{\text{past}}$ and the agent's future trajectory (again, from the expert demonstrations) $t\geq k_0$ to be $\bm{X}_{\text{future}}$. At each control time-step, $k$, the agent is provided with contextual information from the environment, such as a frontal camera view $\bm{\Phi} \in \R^{H\times W \times C}$ and a sequence of waypoints $\bm{\omega}$. Combining $\bm{X}_\text{past}$, $\bm{\Phi}$, and $\bm{\omega}$ we have the agent's observation, or simply  $\mathbf{O}\equiv\{\bm{X}_{\text{past}},\bm{\Phi}, \bm{\omega}\}$. 

At each time-step, the agent must take an action $\mathbf{u}_k$, defined as a tuple of \textit{braking}, \textit{throttling}, and \textit{steering} control. Our objective is to learn a parameterised policy $\pi_\theta$ that maps observations to actions $\mathbf{u} \sim \pi_\theta(\cdot|\mathbf{O})$, such that, given a sequence of observations, an agent that begins at some initial location in the environment can drive to some destination.

\begin{figure*}[t]
\footnotesize
\centering
\includegraphics[width=\linewidth]{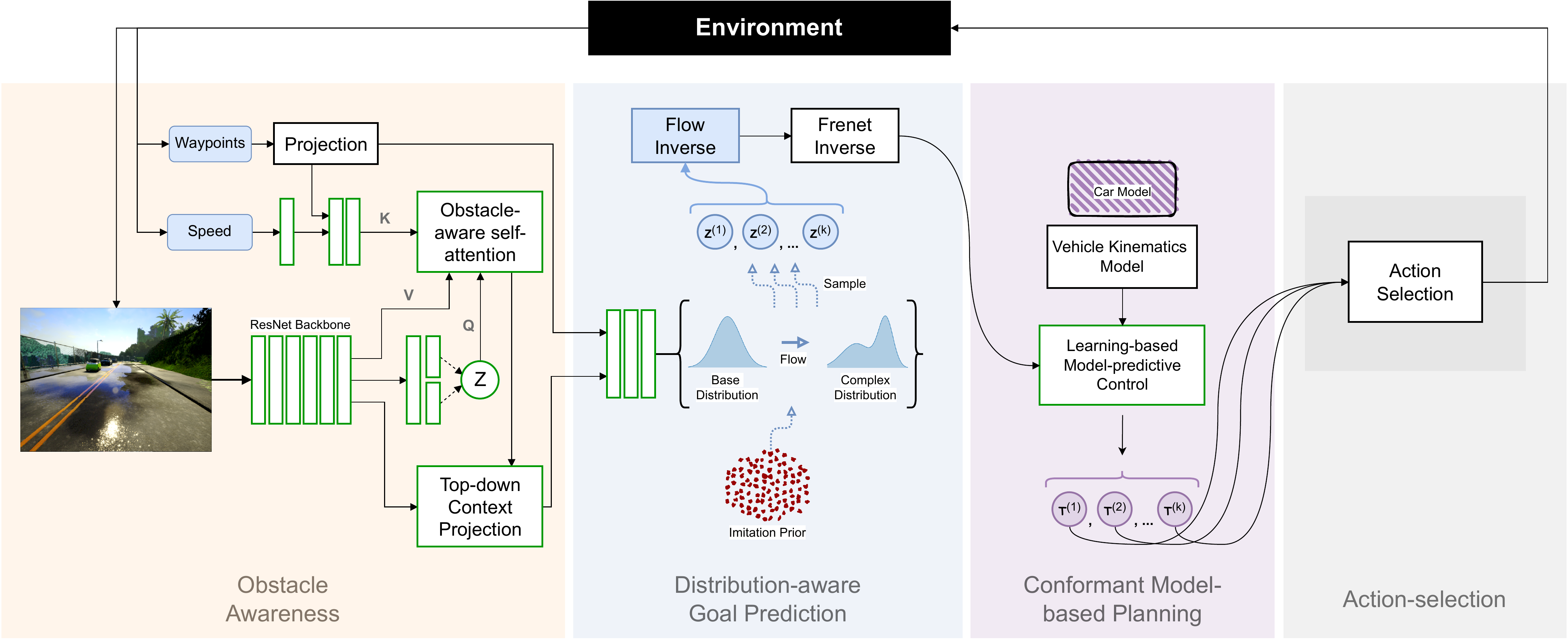}
\caption{Model architecture. Our framework uses the ego-centric sequence of RGB images, world-frame waypoints, and the agent's own current speed information to learn obstacle-aware attention maps and top-down visual representations. These scene encodings inform distribution-aware goal prediction, to leverage expert experience for generalisability to novel scenarios. A set of candidate goal predictions are realised as trajectories, each transformed to the \textit{Frenet} road frame coordinate system and grounded to vehicle kinematics, using a differentiable MPC controller. Trajectories are pruned using a learnable ranking and refinement module.}
\label{ad:l2d:fig:main_figure}
\end{figure*} 


\section{Approach}

Urban driving can be modelled as a composition of driving primitives, where, through decomposition of the conventional multimodal perception backbone into hierarchical units and through modular training, we enjoy lower sample-complexity and improved robustness and generalisability, compared to end-to-end policies.

\subsection{Problem Formulation}

We propose a modular pipeline that performs \textbf{D}istribution-aware \textbf{G}oal \textbf{P}rediction, with conformant model-based planning, in urban driving settings (\ourmodel). It primarily consists of three components: an \textit{obstacle-awareness} module, a \textit{distribution-aware goal prediction} module, and a \textit{conformant model-based planning} module\cut{, and a \textit{trajectory pruning and action-selection} module}, as illustrated in Figures \ref{ad:l2d:fig:issue_with_n2n_imitation} (decomposition) and \ref{ad:l2d:fig:main_figure} (overview). More formally, we factorise the predictive distribution over actions, as a more tractable mapping: \cut{$\mathcal{P} \circ$}$\mathcal{MPC} \circ \mathcal{GP} \circ \mathcal{OA}$. Here, 
$\mathbf{m} \sim \mathcal{OA}(\cdot|\mathbf{O})$ is an obstacle-awareness module, which generates an embedding $\mathbf{m}\in\mathbb{R}^{256}$, given an observation. $\mathbf{x}_{\text{goal}}\in \mathbb{R}^4 \sim \mathcal{GP}(\cdot|\mathbf{m})$ is a goal prediction module, whose samples are desired to be diverse in their coverage of the modes in the true, underlying goal distribution $p^\star(\mathbf{x}_\text{goal}|\mathbf{m})$. Here, ${\mathbf{x}}_\text{goal}$ is the predicted goal from the $\mathcal{GP}$ module and $\mathbf{x}^\star_\text{goal}$ is the true (unobserved) goal of the expert agent, which characterises its scene-conditioned navigational intent. We want $\mathcal{GP}$ to generate multiple samples, where each sample can be regarded as an independent hypothesis of what might have happened, given the same observation. $\mathbf{x}_{k+1:k+N}, \mathbf{u}_{k+1:k+N} = \mathcal{MPC}(\mathbf{x}_{\text{goal}})$ is a learning-based MPC, which takes a predicted goal from the goal distribution as input and generates a trajectory that reaches $\mathbf{x}_{\text{goal}}$ in $N$ time-steps based on its embedded vehicle model. \cut{$\mathcal{P}$ is a pruning module that scores and selects the best trajectory, given an observation $\mathbf{O}$ and a collection of $K$ trajectory candidates.}

Our framework uses the ego-centric sequence of RGB images, world-frame waypoints, and the agent's own current speed information to learn obstacle-aware attention maps and top-down visual representations. These scene encodings inform our goal prediction module, which combines an imitation prior and a goal likelihood objective, in order to leverage expert experience for generalisability to novel scenarios. A set of candidate goal predictions are realised as trajectories, each transformed to the \textit{Frenet} road frame coordinate system and grounded to vehicle kinematics, using a differentiable MPC.\cut{The pruning module scores and filters trajectories, before feeding best trajectories for path-tracking.}

\subsection{Obstacle Awareness: Projections and Encoding}
\label{ad:l2d:ssec:oa}

We condition the learning of our goal distribution on crucial scene context --- from projected topdown feature representation and obstacle self-attention. This perception module's task is to transform the front-view image observations into bird's eye view (BEV) semantic object attention maps.

In this work, we leverage the orthographic feature transform (OFT) technique developed by \cite{roddick2018orthographic}. In particular, we extract obstacle semantic information by pre-training a variational autoencoder \parencite{kingma2014autoencoding} to reconstruct pixels, speed, and steering control in the next time step from current observations. It encourages the latent variables to attend to obstacle in front view (e.g., vehicles, pedestrians, traffic lights, curbs, etc.) which impact future vehicle control.  The front-view feature map $f(u, v)$ is constructed by combining the learned self-attention maps \parencite{vaswani2017attention} with multi-scale images features of pre-trained ResNet-18 front-end. Then, voxel-based features $g(x, y, z)$ are generated by accumulating image-based features $f(u, v)$ to a uniformly spaced 3D lattice $\mathcal{G}$ fixed to the ground plane a distance $y_p$ below the camera and has dimensions $W, H, D$ and a voxel resolution of $r$ using orthogonal transformation. Finally, the topdown image feature representation $h(x, z)$ is generated by collapsing the 3D voxel feature map along the vertical dimension through a learned 1D convolution. In addition to image features, we interpolate waypoint sequence and create a topdown grid representation of waypoints with one-hot encoding. The final topdown feature representation is of dimension $[W/r, D/r, C]$ where the number of channels $C=C_{\text{attn}} + C_{\text{resnet}} + 1$.

\subsection{Multi-mode Goal Distribution}

We wish to approximate the true predictive distribution over all possible goal futures of the ego-agent, $p^\star(\mathbf{x}_\text{goal}| \mathbf{m})$, given the embedding vector $\mathbf{m}$ from the obstacle awareness module (\S \ref{ad:l2d:ssec:oa}) based on the observation $\mathbf{O}$ from the environment. Unfortunately, the predictive \textit{intent} of the expert agent is not observable from the training data: the expert may make multiple manoeuvres given the same scenario, but we only observe the expert taking one of the options.  As a proxy, we take a future state of the expert agent, at fixed time horizon $N$, to be the ``ground-truth" ego-agent's goal, $\mathbf{x}_{\text{goal}}\equiv \mathbf{x}_{k_0+N}$, where N denotes the number of time-steps in the planning horizon. 

Next, rather than learning a mapping to directly imitate these derived expert goals, we instead model an \textit{approximation} $p_\theta(\mathbf{x}_\text{goal}|\mathbf{m})$ of the underlying goal distribution, by leveraging a bijective and differentiable mapping between a chosen \textit{base distribution} $p_0$ and the aforementioned target approximate goal distribution $p_\theta$. This technique is commonly referred to as a `normalizing flow', which provides a general framework for transforming a simple probability density (base distribution) into a more expressive one, through a series of invertible mappings \parencite{tabak2010density, rezende2015variational, papamakarios2021normalizing, kingma2018glow, park2020diverse}.

Formally, let $f$ be an invertible and smooth function, with $f : \R^{d} \rightarrow \R^{d}$, $\mathbf{x}=f(\mathbf{z})$, $\mathbf{z} \sim p_{\mathbf{z}}$, $f^{-1} = g$, and thus $g \circ f(\mathbf{z}) = \mathbf{z}$, for $d$-dimensional random vectors $\mathbf{x}$ and $\mathbf{z}$. In this case, $d=4$ and $p_{\mathbf{z}} = \mathcal{N}(\mathbf{0, \mathbf{I}})$. Further, we attribute to $f$ the property of \textit{diffeomorphism} \parencite{milnor1997topology}, which ensures that $q_{\mathbf{x}}$ remains well-defined and obtainable through a change of variables, and ensures that $p_{\mathbf{z}}$ is uniformly distributed on the same domain as the data space \parencite{liao2021jacobian} --- insofar as both $f$ and its inverse $f^{-1}$ are differentiable and that $\mathbf{z}$ retains the same dimension as $\mathbf{x}$:
\begin{equation}\label{eq:likelihood}
p_{\mathbf{x}}(\mathbf{x}) = p_{\mathbf{z}}(\mathbf{z})\begin{vmatrix}\text{det} \frac{\partial f}{\partial z}\end{vmatrix}^{-1}
\end{equation}
\noindent We can construct arbitrarily complex densities, by \textit{flowing} $\mathbf{z}$ along the path created by a chain of K successive \textit{normalizing} distributions $p_{\mathbf{z}}(\mathbf{z})$, with each successive distribution governed by a diffeomorphic transformation: 
\begin{equation}\label{eq:diffeomorphism}
\mathbf{x} = \mathbf{z}_K = f_K\circ\dots\circ f_2 \circ f_1(\mathbf{z_0})
\end{equation}
Following this sequence of transformations, our main interfaces with the flow-based model are through either sampling or evaluating its density, where, in the former, we sample from $p_{\mathbf{z}}(\mathbf{z})$ and must compute the forward transformation $f$; in the latter, we must compute the inverse transformation $f^{-1}$, its Jacobian determinant, and the $p_{\mathbf{z}}(\mathbf{z})$ density evaluation. The learning objective of the goal prediction model is given by Eqn. \ref{eq:objective}, which is to minimise the KL divergence between the true and approximate goal distribution, or equivalently maximise the log-likelihood of the approximate goal distribution:
\begin{equation}\label{eq:objective}
\begin{aligned}
    \min_\theta \text{KL}(p^\star||p_\theta) &= \max_\theta \mathbb{E}_{x\sim p^\star} \log p_\theta(\mathbf{x})   \\
    &\approx \max_\theta \sum_{x\sim \mathcal{D}} \log p_\theta(\mathbf{x})  
    \end{aligned}
\end{equation}
The likelihood is given in in Eqn. \ref{eq:likelihood}, which is made tractable through the special design of the bijective transformation $f$.  Since we are learning a conditional distribution, we also pass the context embedding $\mathbf{m}$ to the flow model, following prior works \parencite{dinh2016density, papamakarios2017masked}.

\subsection{Grounding Action Predictions with Road Geometry and Vehicle Dynamics}

\begin{figure}[t]
\footnotesize
\centering
\includegraphics[width=0.8\linewidth]{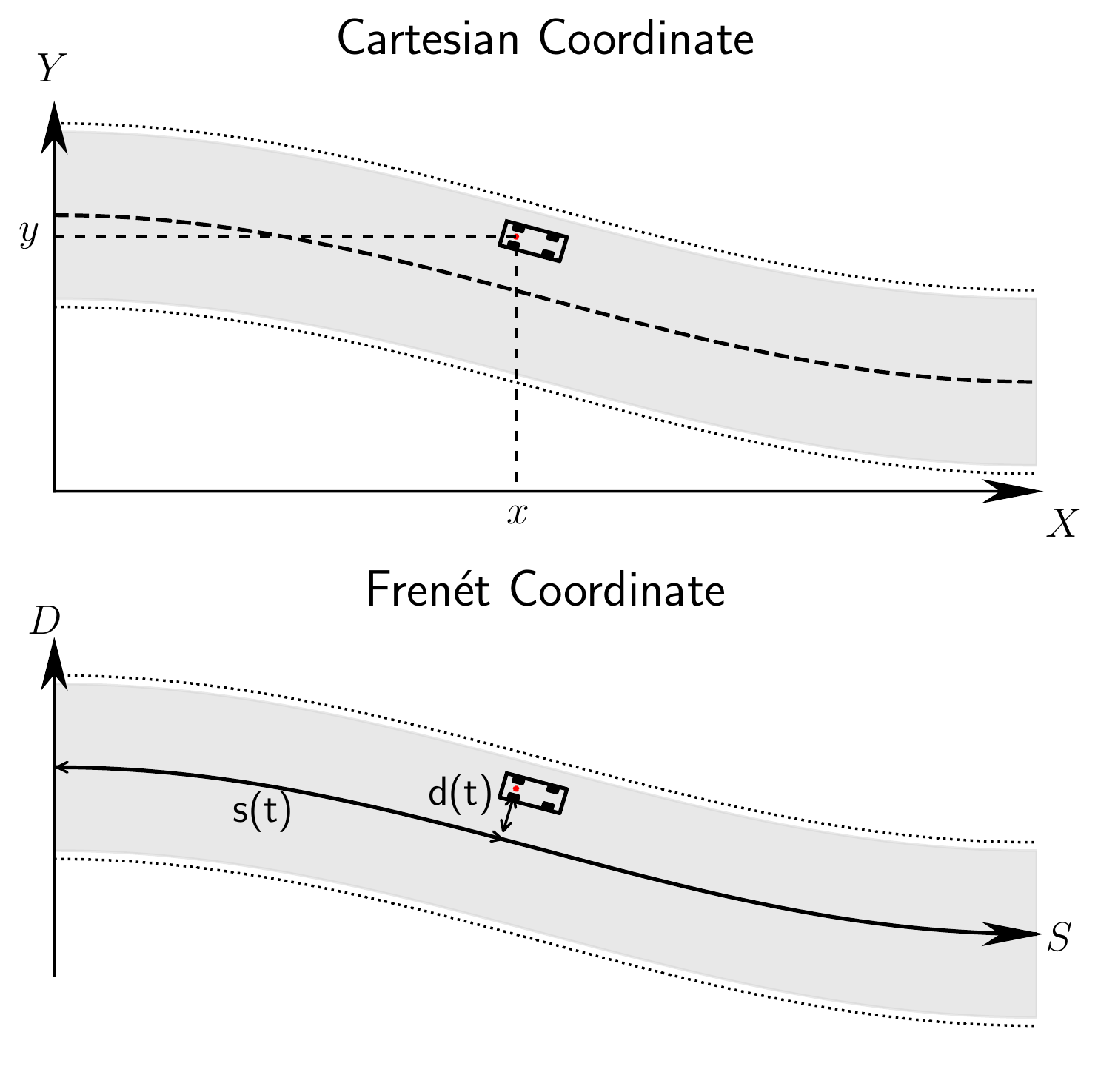}
\caption{Comparison of Cartesian and \frenet~coordinates.}
\label{fig:frenet}
\end{figure}

\begin{figure}[t]
\footnotesize
\centering
\includegraphics[width=0.7\linewidth]{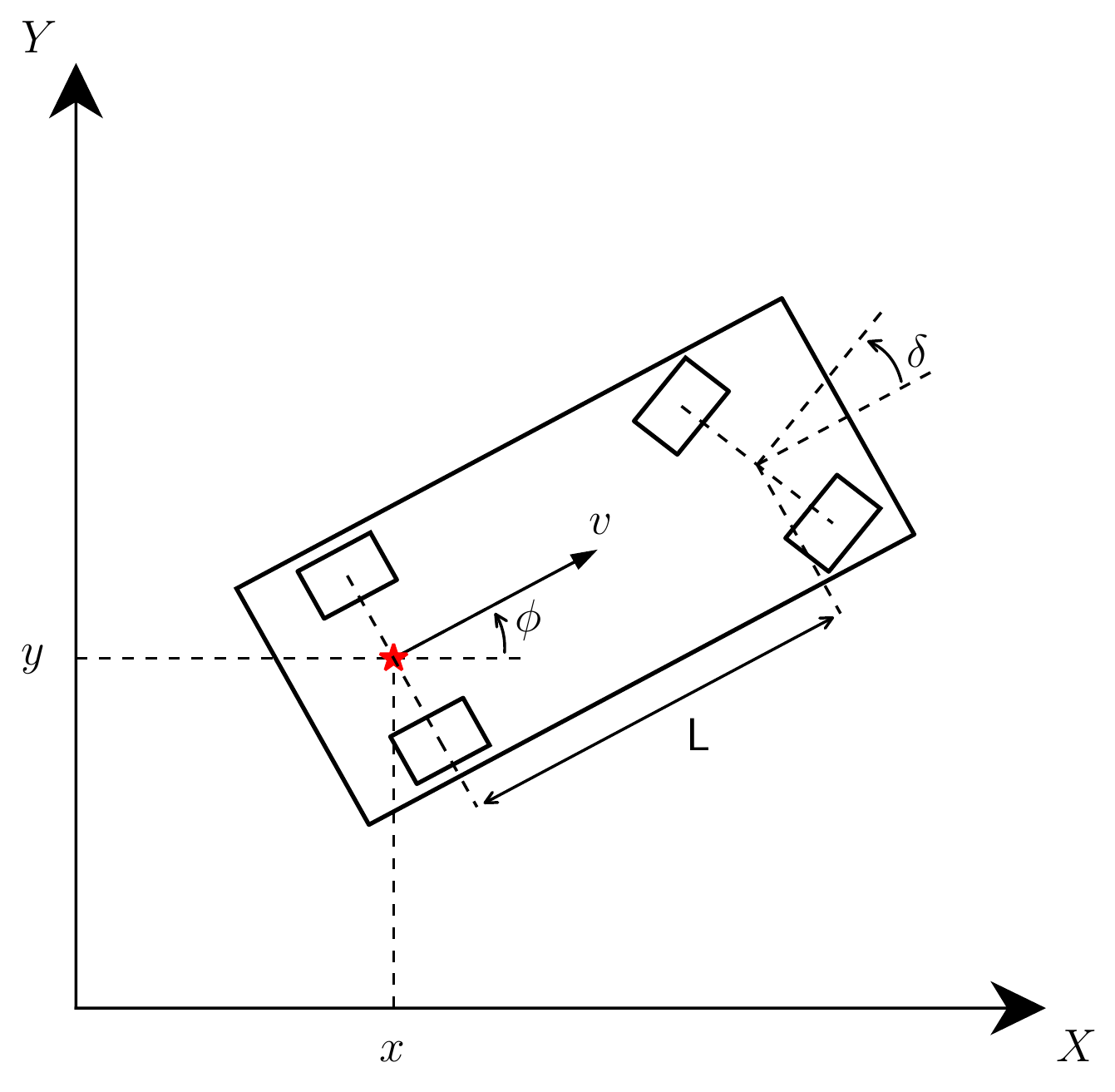}
\caption{Illustration of the kinematic bike model.}
\label{fig:bike_model}
\end{figure}

Given $\mathbf{x}_{\text{goal}}\sim\mathcal{G}\mathcal{P}(\cdot|\mathbf{m})$, we want to find a trajectory to $\mathbf{x}_{\text{goal}}$ that is conformant to both vehicle dynamics and road geometry. To conform to road geometry, we predict the goal state in \frenet~coordinate (Figure \ref{fig:frenet}) and invert to Cartesian coordinate. We parameterise the road geometry as a cubic spline along the way points, and calculate the \frenet~and inverse-\frenet~transformation accordingly. 

To be conformant to the vehicle dynamics, we formulate reaching a desired goal state as a MPC problem (Eqn. \ref{eq:mpc_problem}), which embeds a vehicle model. The objective (Equation \ref{eq:mpc_objective}) is to reach the goal with regularisation on actuations. $\mathbf{Q}$ and $\mathbf{R}$ are both diagonal matrices corresponding to cost weights for tracking reference states and regularising actions. At the same time, the MPC respects the system dynamics of the vehicle (Equation \ref{eq:mpc_sys}), and allowable actuation (Eqn. \ref{eq:mpc_action}). 
\begin{subequations}\label{eq:mpc_problem}
    \begin{align}
    \min_{\mathbf{u}_{1:N}}  \quad & (\mathbf{x}_N-\mathbf{x}_{\text{goal}})^T \mathbf{Q}(\mathbf{x}_N-\mathbf{x}_{\text{goal}}) +\sum_{k=1}^{N}\mathbf{u}_k^T \mathbf{R} \mathbf{u}_k \label{eq:mpc_objective}\\
    \textrm{s.t.} \quad & \mathbf{x}_{k+1} = F(\mathbf{x}_k, \mathbf{u}_k), \quad \forall k = 1, \dots, N \label{eq:mpc_sys}\\
    & \underline{\mathbf{u}}\leq \mathbf{u}_k\leq \bar{\mathbf{u}}\label{eq:mpc_action} 
    \end{align}
\end{subequations}
Specifically, we characterise the vehicle with the kinematic bike model \cite{kong2015kinematic} given in Eqn. \ref{eq:bike_model}\footnote{This set of equations is defined with respect to the back axle of the vehicle.} and visualised in Figure \ref{fig:bike_model}, where the state is $\mathbf{x}=[x, y, v, \phi]$, and the action is $\mathbf{u}=[a, \delta]$.  $a$ is the acceleration, and $\delta$ is the steering angle at the front axle. A key challenge is that the ground truth vehicle parameters were not known to us. Aside from $L$ defined as the distance between the front and rear axle, the kinematic bike model expects actions, i.e. acceleration and steering, in physical units, while the CARLA simulator expects normalised commands. The mapping is unknown to us, and thus, we learn the mapping by minimising the prediction error between predicted and actual trajectories from expert demonstrations.    
\begin{subequations}\label{eq:bike_model}
\begin{align}
\dot{x}&=v\cos(\phi)\\
\dot{y}&=v\sin(\phi)\\
\dot{v}&=a\\
\dot{\phi}&=v\tan{\delta}/L
\end{align}
\end{subequations}


We solve the MPC problem with the iterative linear quadratic regulator (iLQR) proposed in \cite{li2004iterative}, which iteratively linearises the non-linear dynamics along the current estimate of trajectory, solves a linear quadratic regulator (LQR) problem based on the linearized dynamics, and repeats the process until convergence. Specifically, we used the implementation for iLQR from \cite{amos2018differentiable}. 






\section{Experiments}
\label{sec:experiments}

We evaluate our approach using the CARLA \parencite{carla} simulator, with specific focus on three challenge benchmarks: CARNOVEL \parencite{filos2020can}, NoCrash \parencite{codevilla2019exploring}, and the original CARLA benchmark \parencite{carla}. In all settings, the agent is provided with an RGB image, a\cut{high-level command (\textit{NoCrash} and original CARLA benchmarks) or} waypoint sequence\cut{ (CARNOVEL benchmark)}, and vehicle speed; the agent must produce steering, throttle, and braking control, in order to navigation to a destination.

All experiments were conducted using CARLA simulator version 0.9.6, which is worth noting because this version introduced updates to the rendering engine and pedestrian logic, allowing for consistency across \cut{the three} various benchmarks but making the results of contemporary approaches on previous simulator versions no longer comparable \parencite{chen2020learning}. We used a dual-GPU machine, with the following CPU specifications: Intel(R) Core(TM) i9-9920X CPU @ 3.50GHz; 1 CPU,  12 physical cores per CPU, total 24 logical CPU units. The machine includes two NVIDIA Titan RTX GPUs, each with 24GB GPU memory.

In this section, we further describe the tasks, baselines, research challenges, and ablations.

\begin{table*}[t]
\caption{Results of baseline models and our proposed approach on CARNOVEL \parencite{filos2020can}. We report Success Rate ($\uparrow$; M $\times$ N scenes, \%), the Number of Infractions (Collisions, Lane Invasions) per kilometer ($\downarrow$; $\times1e^{-3}$), and Distance travelled (m), on four novel (unseen) scenarios.}
\label{ad:l2d:tab:carnovel_results}
\scriptsize
 \resizebox{\textwidth}{!}{
\centering
\begin{tabular}{p{2.9cm}p{1.9cm}p{1.3cm}p{1.5cm}p{1.9cm}p{1.3cm}p{1.55cm}}
\toprule
 & \multicolumn{3}{c}{\textbf{\textsc{AbnormalTurns}}} &  \multicolumn{3}{c}{\textbf{\textsc{BusyTown}}} \\
\cmidrule(r){2-4} \cmidrule(r){5-7} & Success~($\uparrow$) & Infra/km~($\downarrow$) & Distance & Success~($\uparrow$) & Infra/km~($\downarrow$) & Distance \\
\textbf{Method} & (M $\times$ N scenes, \%) & ($\times1e^{-3}$) & (m) & (M $\times$ N scenes, \%) & ($\times1e^{-3}$) & (m)  \\
\midrule
{CIL \parencite{codevilla2018end}} & {65.71} {\grey{$\pm$ 7.37}} & {7.04} {\grey{$\pm$ 5.07}} & {128} {\grey{$\pm$ 20}} & {5.45} {\grey{$\pm$ 6.35}} & {11.49} {\grey{$\pm$ 3.66}} & {217} {\grey{$\pm$ 33}} \\
{LBC \parencite{chen2020learning}} & {0.0} {\grey{$\pm$ 0.0}} & {5.81} {\grey{$\pm$ 0.58}} & {208} {\grey{$\pm$ 4}} & {20.00} {\grey{$\pm$ 13.48}} & {3.96} {\grey{$\pm$ 0.15}} & {374} {\grey{$\pm$ 16}} \\
{DIM \parencite{rhinehart2018deep}} & {74.28} {\grey{$\pm$ 11.26}} & {5.56} {\grey{$\pm$ 4.06}} & {108} {\grey{$\pm$ 17}} & {47.13} {\grey{$\pm$ 14.54}} & {8.47} {\grey{$\pm$ 5.22}} & {175} {\grey{$\pm$ 26}} \\
{RIP \parencite{filos2020can}} & \textbf{87.14} {\grey{$\pm$ 14.20}} & {4.91} {\grey{$\pm$ 3.60}} & {102} {\grey{$\pm$ 21}} & {62.72} {\grey{$\pm$ 5.16}} & {3.17} {\grey{$\pm$ 2.04}} & {167} {\grey{$\pm$ 21}} \\
\midrule
{\ourmodel~(\textit{ours})} & {82.86} {\grey{$\pm$ 6.39}} & \textbf{1.49} {\grey{$\pm$ 1.44}} & {108.74} {\grey{$\pm$ 8.70}} & \textbf{76.36} {\grey{$\pm$ 4.98}} & \textbf{5.51} {\grey{$\pm$ 2.34}} & {126.83} {\grey{$\pm$ 6.73}} \\
\bottomrule
\toprule
& \multicolumn{3}{c}{\textbf{\textsc{Hills}}} &  \multicolumn{3}{c}{\textbf{\textsc{Roundabouts}}} \\
\cmidrule(r){2-4} \cmidrule(r){5-7} & Success~($\uparrow$) & Infra/km~($\downarrow$) & Distance & Success~($\uparrow$) & Infra/km~($\downarrow$) & Distance \\
\textbf{Method} & (M $\times$ N scenes, \%) & ($\times1e^{-3}$) & (m) & (M $\times$ N scenes, \%) & ($\times1e^{-3}$) & (m)  \\
\midrule
{CIL \parencite{codevilla2018end}} & {60.00} {\grey{$\pm$ 29.34}} & {4.74} {\grey{$\pm$ 3.02}} & {219} {\grey{$\pm$ 34}} & {20.00} {\grey{$\pm$ 0.0}} & {3.60} {\grey{$\pm$ 3.23}} & {269} {\grey{$\pm$ 21}} \\
{LBC \parencite{chen2020learning}} & {50.00} {\grey{$\pm$ 0.0}} & {1.61} {\grey{$\pm$ 0.15}} & {514} {\grey{$\pm$ 101}} & {8.00} {\grey{$\pm$ 10.95}} & {3.70} {\grey{$\pm$ 0.72}} & {323} {\grey{$\pm$ 43}} \\
{DIM \parencite{rhinehart2018deep}} & {70.00} {\grey{$\pm$ 10.54}} & {6.87} {\grey{$\pm$ 4.09}} & {195} {\grey{$\pm$ 12}} & {20.00} {\grey{$\pm$ 9.42}} & {6.19} {\grey{$\pm$ 4.73}} & {240} {\grey{$\pm$ 44}} \\
{RIP \parencite{filos2020can}} & {87.50} {\grey{$\pm$ 13.17}} & {1.83} {\grey{$\pm$ 1.73}} & {191} {\grey{$\pm$ 6}} & {42.00} {\grey{$\pm$ 6.32}} & \textbf{4.32} {\grey{$\pm$ 1.91}} & {217} {\grey{$\pm$ 30}} \\
\midrule
{\ourmodel~(\textit{ours})} & \textbf{100} {\grey{$\pm$ 0.0}} & \textbf{0.0} {\grey{$\pm$ 0.0}} &  {193.12} {\grey{$\pm$ 0.03}} & \textbf{80.00} {\grey{$\pm$ 14.14}} & {4.44} {\grey{$\pm$ 7.21}} & {126.32} {\grey{$\pm$ 11.83}} \\
\bottomrule
\end{tabular}
}
\end{table*}

\para{Benchmark tasks.} We follow \textcite{filos2020can} in assessing the robustness of modelling approaches to novel, OOD driving scenarios in their CARNOVEL benchmark. Predicated on the CARLA simulator \parencite{carla}, agents are first trained on the provided offline expert demonstration from \texttt{Town01} that were originally generated using a rules-based autopilot. Agents are then evaluated on various OOD navigation tasks, such as: abnormal turns, busy-town settings, hills, and roundabouts. Agents' performance are measured according to the following metrics: success rate (percentage of successful navigations to the destination), infractions per kilometre (ratio of moving violations to kilometre driven), and total distance travelled. 



\para{Baselines models.} We compare our model with the following baselines in the CARNOVEL benchmark:

\textit{Conditional imitation learning} (CIL) \parencite{codevilla2018end} is an end-to-end behaviour cloning approach, which conditions its predictions on high-level commands and LiDAR information.

\textit{Learning by cheating} (LBC) \parencite{chen2020learning} extends CIL through cross-modal knowledge distillation, from a teacher network (trained on privileged information --- e.g., environment state, overhead images) to a sensorimotor navigation agent (the ego-agent).

\textit{Deep Imitative Model} (DIM) \parencite{rhinehart2018deep} is a trajectory forecasting and control method, which combines an imitative objective with goal-directed planning.

\textit{Robust Imitative Planning} (RIP) \parencite{filos2020can} is the method that was proposed alongside the recent CARNOVEL benchmark. RIP is an epistemic uncertainty-aware method, targeted toward robustness to OOD driving scenarios.

\section{Results}
\label{sec:results}

\begin{figure*}[t]
\footnotesize
\centering
\includegraphics[width=0.95\linewidth]{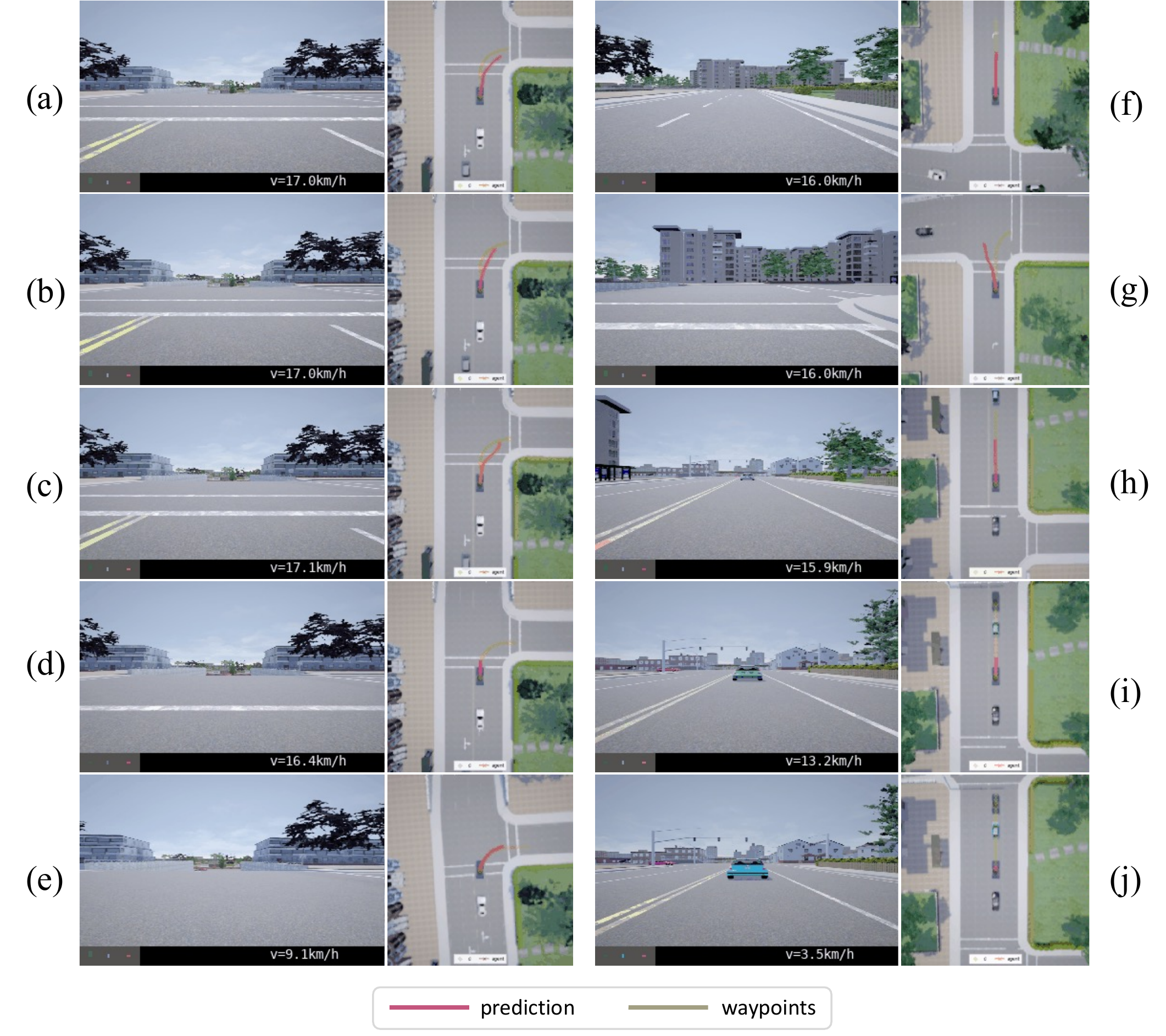}
\caption{Qualitative results from simulated task execution.}
\label{ad:l2d:fig:frames}
\end{figure*}

\parai{Generalisation to OOD scenes.} In Table \ref{ad:l2d:tab:carnovel_results}, we report the success rate, number of infractions per kilometer, and distance travelled of our approach, alongside strong baselines from both the learning-to-drive and trajectory forecasting communities. We show significant improvements in unseen generalisation in visuomotor control for urban driving from our approach, which jointly estimates the ego-agent's action distribution while learning to predict and score intermediate goals. Notably, we observe most significant improvements over the next-best model, Robust Imitative Planning (RIP; \citet{filos2020can}; an epistemic uncertainty-aware model) when transferring models to completely unseen traffic layouts, such as Roundabouts. 

\parai{Qualitative results on agent intention.} A notable benefit of our factorising the urban driving problem---into encoding, distribution-aware goal prediction, conformant model-based planning, and pruning + action-selection---is that we obtain increased interpretability in our agent's prediction pipeline. Specifically, by way of our goal-prediction module, we obtain the ability to reveal agent's intent, during its simulated task execution. Figure \ref{ad:l2d:fig:frames} offers qualitative results from our agent's interaction with the environment during inference. Each frame captures the agent's prediction (in red), given its own speed information, ego-vision context, and short waypoint sequences (in mustard). 

We observed that waypoint sequences, provided by the environment, may sometimes change, as agents approach locations where some decision must be made (e.g., turns). Coupled with our models estimation of the underlying action distribution, conditioned on offline dataset samples, our model correctly makes multiple admissible predictions of turning (a), going straight (b), changing lanes (c), or stopping/slowing (d) as it approaches the first intersection. After committing and executing the turn (e-f), the agent once again considers multiple possible futures (g), before once again following the rightward arcing waypoint sequence. On straight sections, the agent shows strong belief on forward movement, indicated by distant and straight goal predictions (h), but correctly slows (i) and stops (j) for obstacles. 


\section{Discussion and Conclusion}
\label{sec:conclusion}

In this paper, we introduce a distribution-aware trajectory generation mechanism that remains conformant to both road geometry and vehicle kinematics. We show how our agent uses this learned prior to generalise to completely out-of-distribution driving scenarios, such as busy towns, abnormal turns, hills, and unseen traffic patterns such as roundabouts. 



\clearpage 
{
\bibliography{root}
\bibliographystyle{icml2022}
}



\end{document}